%% file: main.tex
\documentclass[conference]{IEEEtran}
\IEEEoverridecommandlockouts
\usepackage{cite}
\usepackage{graphicx}
\usepackage{amsmath,amssymb,amsfonts}
\usepackage{algorithm}
\usepackage{algorithmicx}
\usepackage{algpseudocode}
\usepackage{booktabs}
\usepackage{arydshln}
\usepackage{textcomp}
\usepackage{xcolor}
\def\BibTeX{{\rm B\kern-.05em{\sc i\kern-.025em b}\kern-.08em
    T\kern-.1667em\lower.7ex\hbox{E}\kern-.125emX}}

\newcommand{\eg}{\textit{e.g.~}}
\newcommand{\ie}{\textit{i.e.~}}
\newcommand{\figref}[1]{Fig. \ref{#1}}
\newcommand{\tabref}[1]{Table \ref{#1}}
\begin{document}

\title{
Federated Learning for Chronic Obstructive Pulmonary Disease Classification with Partial Personalized Attention Mechanism
\thanks{\textsuperscript{*}Corresponding Author. Email: chenweidao163@163.com.}
}

\author{
\IEEEauthorblockN{Yiqing Shen\textsuperscript{1,2}, Baiyun Liu\textsuperscript{2}, Ruize Yu\textsuperscript{2}, Yudong Wang\textsuperscript{2}, Shaokang Wang\textsuperscript{2}, Jiangfen Wu\textsuperscript{2}, Weidao Chen\textsuperscript{2*}}
\IEEEauthorblockA{
\textsuperscript{1}Johns Hopkins University, MD, USA 
\textsuperscript{2}Beijing Infervision Technology Co., Ltd., Beijing, China
}
}

\maketitle

\input{1-abstract}
\input{2-introduction}

\input{3-method}
\input{4-result}

\input{5-conclusion}
\bibliographystyle{plain}
\bibliography{main}

\end{document}

%% file: 1-abstract.tex
\begin{abstract}
Chronic Obstructive Pulmonary Disease (COPD) is the fourth leading cause of death worldwide. Yet, COPD diagnosis heavily relies on spirometric examination as well as functional airway limitation, which may cause a considerable portion of COPD patients underdiagnosed especially at the early stage. Recent advance in deep learning (DL) has shown their promising potential in COPD identification from CT images. However, with heterogeneous syndromes and distinct phenotypes, DL models trained with CTs from one data center fail to generalize on images from another center. Due to privacy regularizations, a collaboration of distributed CT images into one centralized center is not feasible. Federated learning (FL) approaches enable us to train with distributed private data. Yet, routine FL solutions suffer from performance degradation in the case where COPD CTs are not independent and identically distributed (Non-IID). To address this issue, we propose a novel personalized federated learning (PFL) method based on vision transformer (ViT) for distributed and heterogeneous COPD CTs. To be more specific, we partially personalize some heads in multiheaded self-attention layers to learn the personalized attention for local data and retain the other heads shared to extract the common attention. To the best of our knowledge, this is the first proposal of a PFL framework specifically for ViT to identify COPD. Our evaluation of a dataset set curated from six medical centers shows our method outperforms the PFL approaches for convolutional neural networks. 
\end{abstract}


%% file: 2-introduction.tex
\section{Introduction}

Chronic Obstructive Pulmonary Disease (COPD) is a prevalent chronic pulmonary condition that remains a major cause of morbidity and mortality worldwide \cite{copd_death,copd_death2}. Present-day spirometric criteria often result in under-diagnosis and under-report in routine exams as COPD patients can be functional intact at their early state \cite{copd_diagnose}. Previous studies have demonstrated the successful application of deep learning (DL) models in the identification of COPD using computed tomography (CT) images as inputs, particularly the convolutional neural networks (CNNs) \cite{copd_dl2,copd_dl}. These models were mostly trained and validated on the large public database and the performance of which in the real-world testing varied depending on the patient source and distribution. Therefore, there is a clear need to generate a universally effective approach for COPD diagnosis in real clinical settings. 

It is well recognized that COPD is a heterogeneous syndrome and adopts distinct phenotypes such as emphysema and airway loss, depending on the genetic variants and inflammatory subtypes \cite{copd_multi}. Thus, a rich source of data covering these phenotypes is critical for understanding the underlying mechanism of COPD heterogeneity. Yet, current privacy regulations \eg \cite{hippa,gdpr} restrict a collaboration of the distributed data into a centralized center, because the availability of such confidential data is not permitted. To narrow this gap and mitigate the risk of privacy leakage, federated learning (FL) \cite{fl}, as a machine learning paradigm, is proposed, where deep learning models are trained locally at the involved client without explicit data communication. Specifically, the clients can keep their data private, and only share weights and gradients for the model update. In this scenario, FL has already shown empirical success in various tasks of medical imaging \cite{fl_med}, \eg COVID-19 CT detection \cite{fl_covid2,fl_covid1}, histology tissue subtype classification \cite{fl_histolgy}. 

Data heterogeneity in COPD CTs, characterized by varying geographical locations, patient distribution, image scanners, and diagnosis protocols, usually hinders the successful deployment of the FL model. Concretely, vanilla federate learning approaches are prone to suffer from performance degradation with non-IID (not independently and identically distributed) data, primarily caused by the weight divergence \cite{fl_noniid}. Subsequently, personalized federated learning (PFL) is introduced to enable local models to learn local personalized weights \cite{pfl}, where the global shared model is personalized for each local client as an intermediate paradigm between pure local training and FL. Model weights personalization has received notable attention \cite{fedper,lgfed}. For example, \textproc{FedPer} decouples a neural network architecture into global base layers to be trained by \textproc{FedAvg}, and the top personalized layers to address the label distribution skew \cite{openchallenge}. \textproc{LG-FedAvg} is proposed by personalizing the bottom layer to tackle the feature distribution skew \cite{lgfed}. Yet, these PFL approaches are primarily restricted to convolutional neural network (CNN) structures, without any specific design for Vision Transformer (ViT). In this paper, we first develop a PFL model for ViT \cite{vit} because of its potential to tackle the heterogeneity in medical images \cite{why_vit}. Innovatively, a partial personalization scheme is proposed for the Multiheaded Self-Attention (MSA) layers in ViT.

The major contributions are four-fold. First, we propose an FL model based on ViT for COPD CT image classification. Second, to address the data heterogeneity, a novel partial personalization for the multiheaded self-attention layer in ViT is developed. To the best of our knowledge, this is the first attempt at a PFL scheme for ViT. Third, to facilitate the PFL training, we propose a consistent regularization between the global shared and local private attention. Finally, experimental results on COPD data collected from multiple medical centers yield the effectiveness of our approach in comparison with the CNN-based PFL methods.

%% file: 3-method.tex
\begin{figure*}[htbp!]
    \centering
\includegraphics[width=0.74\linewidth]{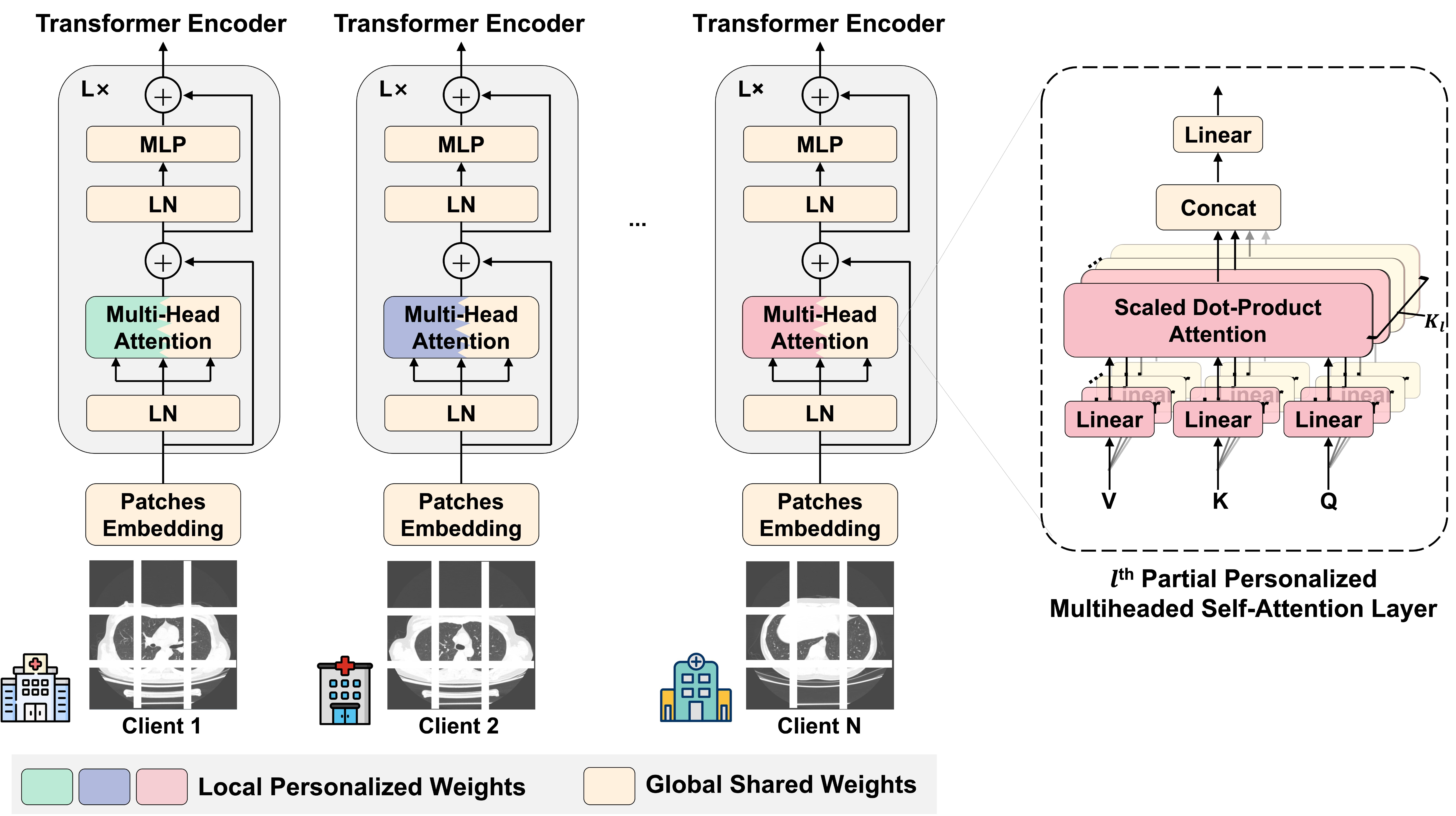}
    \caption{The semantic illustration of partial personalization scheme for multiheaded self-attention layers in ViT. In MSA layers, we partially personalize some heads to learn the attentions for local personalized features, and the rest are remained shared globally to learn common attentions. The personalized weights reside locally, and the shared weights are aggregated globally.}
    \label{fig:personalized_vit}
\end{figure*}

\section{Methods}
In this section, we elaborate on the personalized federated multiple-instance learning framework based on ViT for COPD classification.

\textbf{ViT Backbone.} We decompose the three dimensional CT volume input $\mathbf{x} = [x_1,\cdots,x_n]\in\mathbb{R}^{H\times W \times C}$, where $(H,W)$ is the image resolution, into $C$ number of 2D slices \ie $x_i \in \mathbb{R}^{H\times W}$ ($i=1,\cdots,n$) to perform image flatting. Then, we denote the patch input as $x_i^p \in \mathbb{R}^{N\times P^2}$ for slice $x_i$, where $(P, P)$ is the patch resolution and $N = HW/P^2$ is the number of patches in one slice. Afterwards, a learnable embedding is employed to vectorize the patch inputs into vector embedding $\mathbf{z}_0$ \cite{vit}. Rather than the convolutional layers, Vision Transformer (ViT) \cite{vit} adopts Multi-head Self-Attention (MSA) and Multi-Layer Perceptron (MLP) operators on the embeddings. To be more specific, the latent output of $l$\textsuperscript{th} layer, for $l=1,\cdots,L$, $\mathbf{z}_l$ is computed as: 
\begin{equation}
\begin{aligned}
    \mathbf{z}_l^\prime & = \operatorname{MSA}_l \left(\operatorname{LN}_{l,0}(\mathbf{z}_{l-1})\right) + \mathbf{z}_{l-1}, \\
    \mathbf{z}_l & =  \operatorname{MLP}_l\left(\operatorname{LN}_{l,1}(\mathbf{z}_{l-1}^\prime)\right) + \mathbf{z}_{l-1}^\prime\,,
\end{aligned}   
\end{equation}
where we use the superscript $l$ to denote the weights associated with the $l$\textsuperscript{th} layer, $\operatorname{LN}(\cdot)$ represents the layernorm \cite{layernorm}. 

\textbf{Partial Personalization Scheme for MSA Layer in FL.} We propose a model personalization scheme for ViT in FL by keeping a designed percentage of weights in the attention layers residing locally during FL training, while the rest are shared across all clients. Specifically, a proportion of MSAs is personalized in each layer to learn local personalized attention representations from heterogeneous CT data, while the rest parameters retains globally shared to extract the common attention representations. We write the $l$\textsuperscript{th} MSA layer as
\begin{equation}
    \begin{aligned}      [\mathbf{q}_{l,k},\mathbf{k}_{l,k},\mathbf{v}_{l,k}] & = \mathbf{z}_{l-1} \mathbf{U}_{\mathbf{qkv},l,k},~~\text{for}~~k=1,\cdots,K_l; \\
        SA_{l,k}(\mathbf{z}_{l-1}) &= \operatorname{softmax}(\mathbf{q}_{l,k} \mathbf{k}_{l,k}^T/\sqrt{D_h})\mathbf{v}_{l,k} \\
        MSA_{l}(\mathbf{z}_l) & = [SA_{l,1}(\mathbf{z}_{l-1}),\cdots,SA_{l,K_l}(\mathbf{z}_{l-1})]\mathbf{U}_{\mathbf{msa},l}
    \end{aligned}
\end{equation}
where $D_h$ is the size of vectors $\mathbf{q}_{l,k},\mathbf{k}_{l,k},\mathbf{v}_{l,k}$, $K_l$ is the total number of self-attention heads of $\mathbf{U}_{\mathbf{msa},l} = [\mathbf{U}_{\mathbf{msa},l,1},\cdots \mathbf{U}_{\mathbf{msa},l,K_l}]^T$. We follow the same terminology and definitions of $\mathbf{U}$ as in \cite{vit}. To enable the ViT to capture the local personalized attention for heterogeneous data, we personalize a number of $P_l < K_l$ heads in the MSA layers. Specifically, the collection of personalized weights in $l$\textsuperscript{th} MSA comprises two 
parts \ie 
\begin{align}
\mathbf{U}_{\mathbf{qkv},l}^p &= [\mathbf{U}_{\mathbf{qkv},l,1},\cdots,\mathbf{U}_{\mathbf{qkv},l,P_l}]\\
\mathbf{U}_{\mathbf{msa},l}^p &= [\mathbf{U}_{\mathbf{msa},l,1},\cdots,\mathbf{U}_{\mathbf{msa},l,P_l}],
\end{align}
where superscript $p$ denotes the personalized weights. On the other hand, the global shared weights are made up of 
$\mathbf{U}_{\mathbf{qkv},l}^g = [\mathbf{U}_{\mathbf{qkv},l,P_l+1},\cdots,\mathbf{U}_{\mathbf{qkv},l,K_l}]$ and $\mathbf{U}_{\mathbf{msa},l}^g = [\mathbf{U}_{\mathbf{msa},l,P_l+1},\cdots,\mathbf{U}_{\mathbf{msa},P_l+1,K_l}]$,
where superscript $p$ denotes the personalized weights. The global shared weights will be uploaded to the central server and aggregated by \textproc{FedAvg} in the training stage \cite{fl} to extract the global shared attentions. For model simplicity, we use a uniform personalization ratio for all MSAs \ie $\frac{P_l}{K_l} = p\in(0,1)$ for $l =1,\cdots L$. Then, the value of $p$ determines the strength of model personalization. For example, $p=1$ degrades the method to be a local training way i.e., no shared parameters via the central server except the MLP layers; while $p=0$ becomes the vanilla \textproc{FedAvg}. The overall architecture of our proposed partial personalization scheme for attention layers is depicted in \figref{fig:personalized_vit}.

\textbf{Consistency Regularization between Local Personalized and Global Shared Weights.} The partial personalization scheme enables us to impose an additional regularization on the local personalized weights. We term the globally shared subnet in ViT excluding all personalized MSA weights \ie $\{\mathbf{U}_{\mathbf{qkv},l}^p,\mathbf{U}_{\mathbf{msa},l}^p|l=1,\cdots,L\}$, as $\operatorname{ViT}^g(\cdot)$ \cite{inplacekd}. In parallel, we name another client-specific subnet with only personalized MSA weights as $\operatorname{ViT}^{p,j}(\cdot)$ where $j$ is the associated client ID. We impose a consistency regularization to force the local personalized weights and shared weights ultimately generate predictions that coincide with each other. To achieve this goal, we propose a mutual consistent regularization, depicted as follows
\begin{equation}
\begin{aligned}
    \mathcal{L}_{con}^j &= \mathbf{KL}(
    \sigma(\operatorname{ViT}^g(\mathbf{x})/T), 
    \sigma(\operatorname{ViT}^{p,j}(\mathbf{x})/T)
    )
    + \\
    &\mathbf{KL}(
    \sigma(\operatorname{ViT}^{p,j}(\mathbf{x})/T),
    \sigma(\operatorname{ViT}^{g}(\mathbf{x})/T)
    )\,.\label{eq:regularizartion}
\end{aligned}
\end{equation}
The softmax $\sigma(\cdot)$ softens the outputs of subnets after being divided by the temperature $T$. KL-divergence $\mathbf{KL}(\cdot,\cdot)$ is not interchangeable, hence we symmetry it in Eq. \eqref{eq:regularizartion}. In the local training stage at $j$\textsuperscript{th}, the overall objective is 
\begin{align}
    \mathcal{L}^j = &\mathcal{L}_{ce} +\lambda \cdot \mathcal{L}_{con}^j , ~~\text{with} \label{eq:objective}\\
    &\mathcal{L}_{ce} = \operatorname{CrossEntropy}(\operatorname{ViT}(\mathbf{x}),y),
\end{align}
where $y$ is the ground truth associated with input image $\mathbf{x}$. Coefficient $\lambda$ balances the cross entropy loss and the consistency regularization term.

%% file: 4-result.tex
\section{Experiments}

\begin{figure}[h!]
    \centering
    \includegraphics[width=\linewidth]{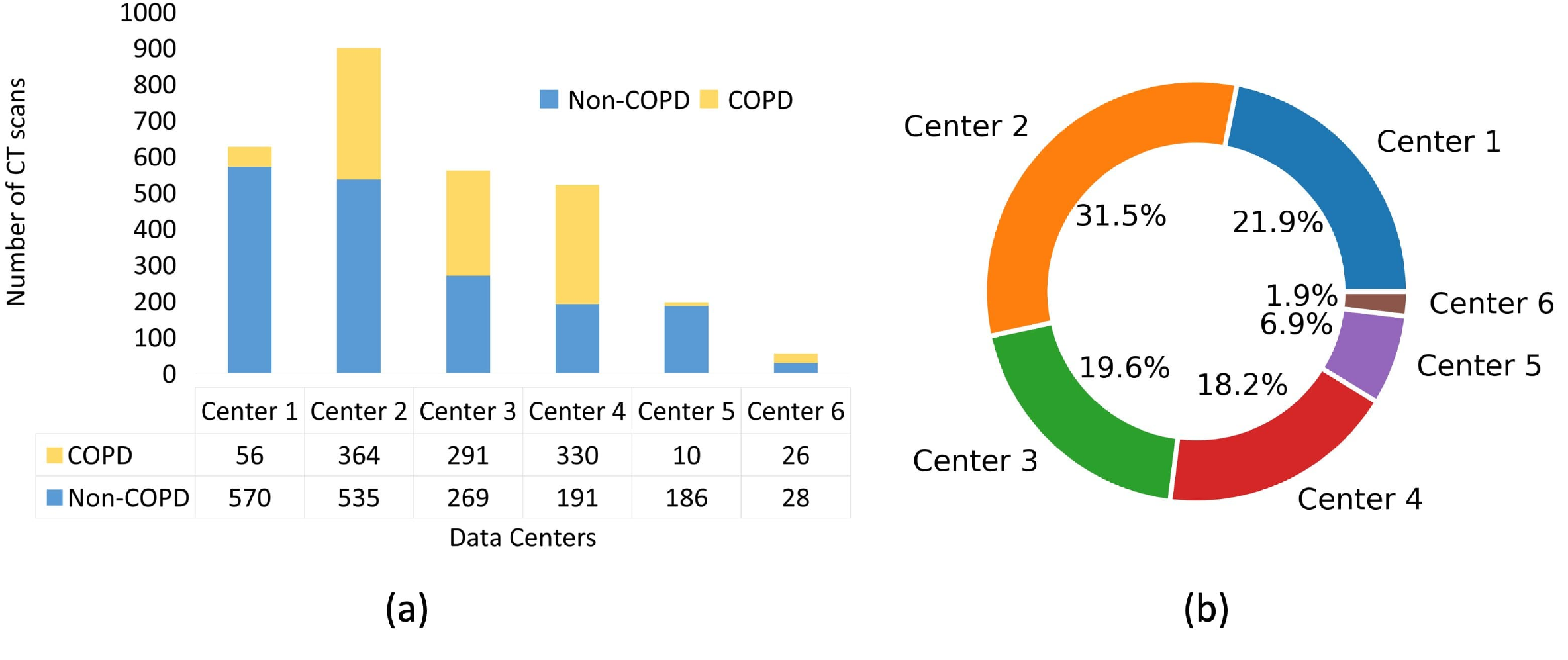}
    \caption{Data scale (a) and the distributions (b) of COPD CT slides from six medical centers used in our evaluation.}
    \label{fig:dataset_percentage}
\end{figure}

\textbf{Datasets.} This study retrospectively collects and utilizes CT images of large patient cohorts from a public dataset NLST (Center 1) \cite{copd_dataset_public} and five hospitals (Center 2-6) \ie a total number of six medical centers. Post-bronchodilator pulmonary function testing was available for all patients and applied as the primary criteria for COPD diagnosis \ie a forced expiratory volume in 1 sec (FEV1)/forced vital capacity (FVC) ratio of $<$ 0.7. The inclusion criteria were (i) participants that underwent single inspiratory breath-hold CT scans in the supine position; (ii) at least one-time pulmonary function test; and (iii) aged 18 years and older and with no history of thoracic surgery. The exclusion criteria were (i) incomplete clinical or radiological data; (ii) compromised image quality; and (iii) substandard spirometric data. Data from different centers were collected from similar periods to exclude the selection bias on the time scale. The NLST public dataset included participants between 55 and 74 years old, with a smoking history of more than 30 pack-years and no self-reported history of lung cancer, thereby allowing us to include low-dose CT acquired among smokers in the dataset, which further adds to the data heterogeneity. 
A total of 2856 participants were enrolled in the final cohort. Among them, 1077 patients were diagnosed with COPD and the other 1779 patients with non-COPD. The data volume of each dataset and data distributions are depicted in \figref{fig:dataset_percentage}. 
Briefly, contiguous CT images were reconstructed in the transverse (axial) plane using standard reconstruction algorithms, creating three-dimensional volumes with up to 5.0 mm slice thickness.

\textbf{Implementations.} We implemented our methods and the baseline approaches by Pytorch on one NVIDIA Tesla V100 GPU device, where the distributed environment of FL is simulated by training the local epoch sequentially.  
We use a consistent training scheme. A stochastic gradient descent optimizer is adopted for each local model update, where the Nesterov momentum and the weight decay rate are set to 0.9 and $5\times10^{-4}$ respectively. The total federated epoch number is set to 50, with the local epoch number to 3. The temperature $T$ in Eq. \eqref{eq:regularizartion} is set to 4.0, the balancing coefficient in Eq. \eqref{eq:objective} as 1, and the personalization ratio $p=0.6$.

\input{4-1-table}

\textbf{Method Comparison.} We compare our method with the baseline vanilla FL \textproc{FedAvg} and local training. We also compare our proposed method with CNN-based PFL approaches, where ResNet32 and ResNet110 \cite{resnet} are selected as the CNN backbones for their compact structures and outstanding performance in the medical domain. We use ViT-Small \cite{vit} as the backbone in our PFL approach. All backbone networks are trained from scratch, without loading any pre-trained weights for a fair comparison. Two state-of-the-art personalization are compared, namely \textproc{FedPer} \cite{fedper} and \textproc{LG-FedAvg} \cite{lgfed}. For the extra hyper-parameters involved in these compared methods, we retain their original settings. We use the AUC (Area Under Curve) as the evaluation metric in two configurations, namely local test and new test \cite{lgfed}. In the local test, we know precisely the client ID where the CT images come from, thus we can choose the particular associated local model. These metrics intend to measure the performance of local personalized weights. In the new test, the client ID of the test sample is unknown, thus we employ an ensemble of all local models to derive averaged predictions, intending to measure the compatibility between local and global weights.

\begin{figure}[htbp!]
    \centering
\includegraphics[width=0.86\linewidth]{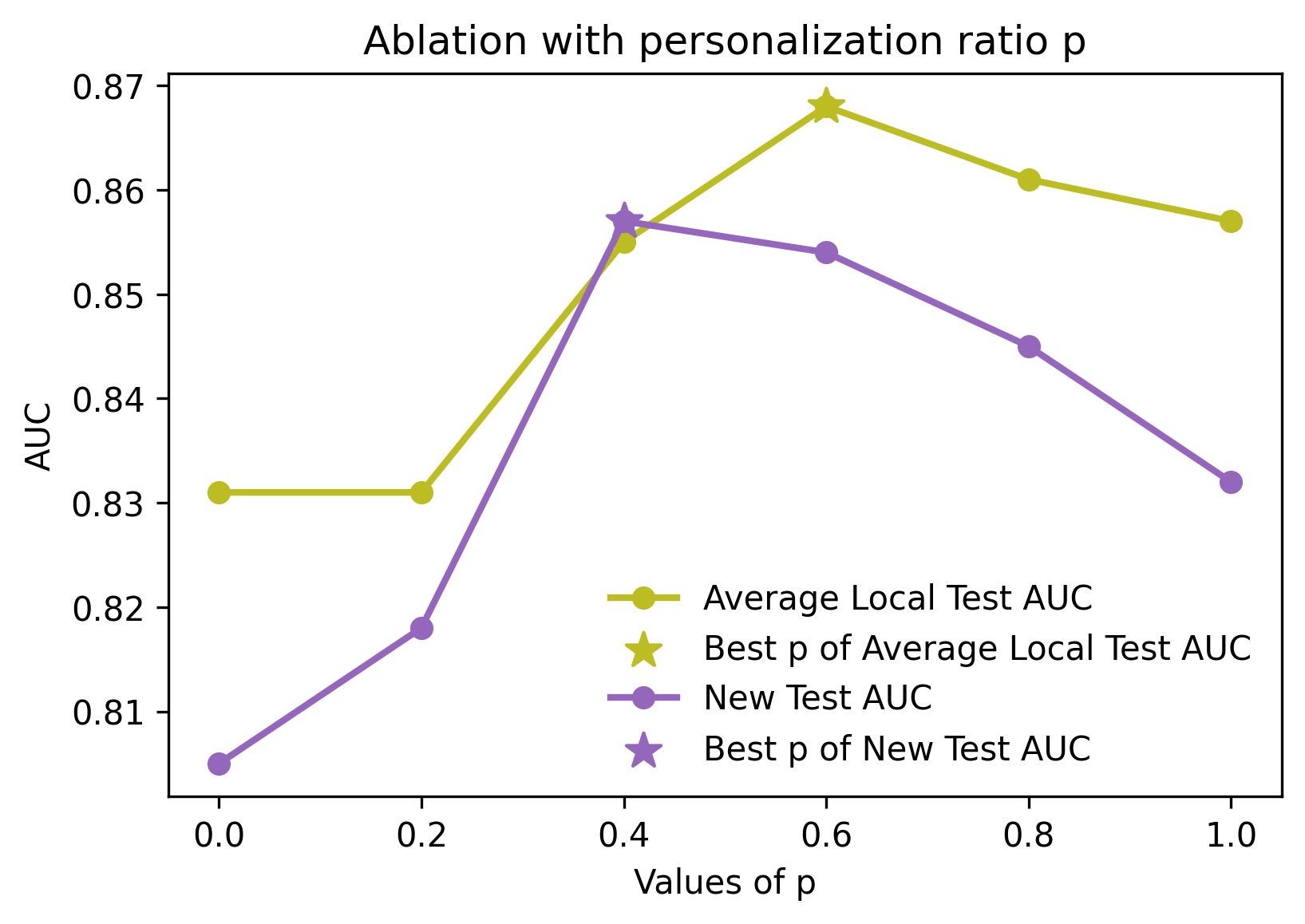}
    \caption{Ablation on the effect of personalization ratio $p$ in terms of averaged local test AUC and new test AUC.}
    \label{fig:ablation}
\end{figure}

\textbf{Results.} As shown in \tabref{tab:result}, our personalization scheme can improve baseline \textproc{FedAvg} by 0.012 to 0.065 local test AUC and 0.049 new test AUC. Moreover, our approach outperforms the state-of-the-art CNN-based PFL methods and illustrates a higher improvement rate to these methods in terms of local test AUC. Our method achieves the highest new test AUC indicating that our model is capable of extracting the common attention with the global shared weights, in addition to the local personalized attention with the personalized weights. With the consistent regularization \ie $\mathcal{L}_{con}$ in Eq. \eqref{eq:regularizartion}, our method achieves extra performance gain, suggesting the effectiveness of each component of our method. 
As shown in \figref{fig:ablation}, the method achieves the highest performance in terms of Local Test AUC when $p=0.6$ and New Test AUC with $p=0.4$,
which implies that a competitive proportion of local personalized knowledge can mitigate the data shift dilemma among different local data centers. 

%% file: 4-1-table.tex
\begin{table}[h!]
\centering
\caption{Performance comparison with local training, FL with various network architectures and PFL with CNNs \cite{fedper,lgfed}. We demonstrate the performance of our approach both with and without the consistent regularization \ie $\mathcal{L}_{con}$. `Local \# j' denotes the local test AUC of the Center j, and `new' denotes the new test AUC. The best performance is marked in \textbf{boldface}, and the second best is \underline{underlined}.}\label{tab:result}
\resizebox{\linewidth}{!}{ 
    \begin{tabular}{l|cccccc|c}
    \toprule
    methods & local\#1 & local\#2 & local\#3 & local\#4 & local\#5 & local\#6 & new \\
    \hline
    Local Train+ResNet32 & 0.720 &  0.709 & 0.674 & 0.688 & 0.612 & 0.613 & 0.674 \\
    Local Train+ResNet110 & 0.731 & 0.720 & 0.675 & 0.692 & 0.631 & 0.618 & 0.680\\
    Local Train+ViT & 0.712 & 0.714 & 0.678 & 0.701 & 0.621 & 0.634 & 0.687 \\
    \hdashline
    \textproc{FedAvg}+ResNet32 & 0.821 & 0.834 & 0.832 & 0.827 & 0.790 & 0.829 & 0.783 \\
    \textproc{FedAvg}+ResNet110 & 0.828 & 0.836 & 0.835 & 0.815 & 0.799 & 0.835 & 0.794 \\
    \textproc{FedAvg}+ViT & 0.835 & 0.839 & 0.841 & 0.842 & 0.811 & 0.831 & 0.805 \\
    \hdashline
    \textproc{FedPer}+ResNet32 & 0.844 &  0.853 & 0.847 & 0.848 & 0.832 & 0.810 & 0.773 \\
    \textproc{FedPer}+ResNet110 & 0.839 & 0.852 & 0.852 & 0.849 & 0.834 & 0.816 & 0.776 \\
    \hdashline
    \textproc{LG-FedAvg}+ResNet32 & 0.867 & 0.859 & 0.856 & 0.841 & 0.845 & 0.837 & 0.802 \\
    \textproc{LG-FedAvg}+ResNet110 & 0.871 & 0.858 & 0.857 & 0.832 & 0.859 & 0.810 & 0.809 \\
    \hdashline
    Ours (w/o $\mathcal{L}_{con}$) & \underline{0.883} & \underline{0.864} & \underline{0.870} & \underline{0.851} & \underline{0.865} & \underline{0.839} & \underline{0.841} \\
    Ours (w $\mathcal{L}_{con}$) & \textbf{0.887} & \textbf{0.871} & \textbf{0.874} & 
    \textbf{0.862} & \textbf{0.876} & \textbf{0.843} & \textbf{0.854} \\
    \bottomrule
    \end{tabular}
}
\end{table}

%% file: 5-conclusion.tex
\section{Conclusion}

In this paper, a novel partial personalization scheme is proposed, specifically for vision transformers in the federated learning paradigm. To be more specific, some heads in the multiheaded self-attention layer are personalized, while the rest heads remain globally shared. The partial personalization scheme intends to learn both local attention and global attention. We evaluate our method on COPD CT images curated from six data centers. Our method, with a consistent regularization between local and global self-attention heads, significantly improves the baseline \textproc{FedAvg} and outperforms other CNN-based PFL methods. One interesting future direction is to evaluate our approach to different data modalities. 